\documentclass{article}
\usepackage{ijcai22}

\usepackage{times}

\usepackage{soul}
\usepackage{url}
\usepackage[hidelinks]{hyperref}
\usepackage[utf8]{inputenc}
\usepackage[small]{caption}
\usepackage{graphicx}
\usepackage{amsmath}
\usepackage{booktabs}
\usepackage{amsmath} 
\usepackage{amssymb}
\usepackage{multirow}
\usepackage{mwe}
\usepackage{subcaption}
\usepackage{tikz}
\usepackage{algorithm}
\usepackage{algorithmic}
\title{Positive-Unlabeled Domain Adaptation}

\author{
Jonas Sonntag$^{1,2}$\footnote{Contact Author}\and
Gunnar Behrens$^2$\And
Lars Schmidt-Thieme$^{1}$\\
\affiliations
$^1$University of Hildesheim\\
$^2$Volkwagen Financial Services\\
\emails
sonntag@uni-hildesheim.de%,gunnar.behrens@vwfs.io,
}

\begin{document}

\maketitle

\begin{abstract}
Domain Adaptation methodologies have shown to  effectively generalize from a labeled source domain to a label scarce target domain. Previous research has either focused on unlabeled domain adaptation without any target supervision or semi-supervised domain adaptation with few labeled target examples per class. On the other hand Positive-Unlabeled (PU-) Learning has attracted increasing interest in the weakly supervised learning literature since in quite some real world applications positive labels are much easier to obtain than negative ones. In this work we are the first to introduce the challenge of Positive-Unlabeled Domain Adaptation where we aim to generalise from a fully labeled source domain to a target domain where only positive and unlabeled data is available. We present a novel two-step learning approach to this problem by firstly identifying reliable positive and negative pseudo-labels in the target domain guided by source domain labels and a positive-unlabeled risk estimator. This enables us to use a standard classifier on the target domain in a second step. We validate our approach by running experiments on benchmark datasets for visual object recognition. Furthermore we propose real world examples for our setting and validate our superior performance on parking occupancy data.   
\end{abstract}

\section{Introduction}
One of the biggest challenge in applied machine learning is the absence of labeled data in real world applications while most deep learning methods rely on supervised training often requiring several thousands of labeled training data. Recently this problem gets more and more acknowledged from the research community as the direction of few shot learning \cite{Wang2020} is gaining attention. \\ 
In this work we will unite two different approaches on few shot learning in one single framework, namely domain adaptation and positive unlabeled learning and we are the first to introduce the problem type and a methodology for the combination of both. The idea of domain adaptation is to transfer knowledge from a label rich source domain to a label scarce target domain. A rich line of research focuses on unlabeled domain adaptation (UDA) \cite{wang2021discriminative}, where the target domain contains no labels at all while some methods consider semi-supervised domain adaptation (SSDA) where few labeled examples per class are available in the target domain. The biggest challenge for all kinds of domain adaptation is to cope with the shift of data distribution between different domains. Most approaches tackle this challenge by learning a domain invariant latent feature representation. \\
Another scenario in few shot learning is Positive-Unlabeled (PU) learning, as a special case of semi-supervised learning. In this binary classification scenario the target domain contains a small number of positive labeled samples and a large amount of unlabeled data containing both positive and negative samples. PU learning arises naturally in many applications where positive labels are much easier to obtain than negative labels. Examples include medical MRI image classification where not yet diagnosed patients can not be treated as negative \cite{Chen2020}. \\
Both PU learning and domain adaptation have been studied intensively, however to the best of our knowledge there is no research yet to the combination of the two, i.e. predicting in a target domain with only few positive labels given a related, fully labeled source domain. 
In this work we are the first to present a learning approach for this problem and validate it's performance on well known benchmark datasets. Moreover we introduce real world problem settings where currently PU learning is used but fully labeled related domains are available and show the superiority of our methodology for parking availability prediction.   
%From the perspective of domain adaptation research our problem setting is between semi-supervised domain adaptation \cite{Saito2018} where the target domain contains few labels of all classes and unsupervised domain adaptation where only the source domain is labeled. From the perspective of PU learning we show that including a completely labeled dataset from a different domain in the training process brings a lift compared to pure PU-learning methods. This is an important note since often labeling negative examples is not generally impossible but simply much more time-consuming and expensive \cite{Hsieh2019}. Hence the option to reuse a labeled dataset for related tasks can make the effort more lucrative. \\
To summarize our contributions are the following
\begin{itemize}
    \item We are the first to introduce the problem setting of Positive-Unlabeled Domain Adaptation. 
    \item We investigate how existing state of the art models from domain adaptation and PU learning perform in this new setting for both real world and benchmark datasets.
    \item We propose a two-step learning methodology designed for this new problem type and consistently outperform the baselines given only a few positive labels in the target domain.
\end{itemize}

\section{Related Work}
Few shot learning is a type of machine learning problem where the algorithm is trained to perform a target task with only limited supervised training data \cite{Wang2020}. 
Semi-supervised learning tackles the challenge of limited labels by leveraging information from unlabeled examples. 
Positive-Unlabeled learning is a special case of semi-supervised learning where only positive labels are given while the unlabeled dataset contains positive as well as negative observations \cite{Bekker2020}. \\
Domain Adaptation on the other hand uses training data from an other related domain where more data is available to transfer knowledge to the label scarce domain. \\
There is no research yet on the combination of the two areas. There is related work dealing with covariate shift in PU-learning \cite{Sakai2019,Hammoudeh2020}. These methods however deal with a pure positive-unlabeled setting where the distribution of positive data shifts between training and testing. They do not consider a fully labeled related source domain and are therefore fundamentally different from our approach. The authors of \cite{Loghmani2020} use PU-learning methodologies to find previously unseen classes in an unlabeled target domain. Although this is an application of PU learning to domain adaptation they do not consider a target domain with only positive labels. \\
The idea of creating pseudo-labels is frequently used in semi-supervised learning \cite{rizve2021,Arazo2020} and is therefore not new in itself. These methods construct the pseudo labels solely based on unlabeled data in the same domain and therefore also miss the domain transfer that our methodology is leveraging.       
\subsection{Positive-Unlabeled Learning}
In many applications positive labels are easier to obtain, whereas representative negative labels are prohibitively expensive or difficult to sample. In the automated diagnosis of diabetes patients one has access to diagnosed patients while non-diagnosed patients still have a high likelihood of being undiagnosed diabetes positive \cite{Claesen2015}. Other areas of interest include land-cover-classification \cite{Li2011}, knowledge base completion \cite{Neelakantan2015} or outlier detection \cite{Blanchard2010}. \\
A common strategy for learning with PU data is to use special unbiased risk estimators \cite{Plessis2015,Kiryo2017} that exploit the fact that the unlabeled data $U$ is a weighted mixture of positive $P$ and negative $N$ data. Another strategy is the so called two-step approach \cite{Bekker2020} where in a first step reliable negative examples are selected and in a second step (semi-) supervised learning methods are applied. Recently \cite{Chen2020} proposed a self learning PU algorithm that combines both ideas by first selecting confident positive and negative examples based on PU-risk and in a latter stage performs self-training with student and teacher models. \\ 
A third approach to tackle PU-problems are Generative Adversarial Networks (GANs) used to generate synthetic positive and negative examples \cite{Hou2018}. 

\subsection{Domain Adaptation}
Domain Adaptation aims at utilizing data from a label-rich source domain to solve a classification problem in a label-scarce target domain. We distinguish homogeneous domain adaptation where the feature space in both domains is the same (e.g. images of same resolution) and heterogeneous domain adaptation where this assumption does not hold and more general transfers (e.g. text to image) can be considered \cite{Yao2019a}. In our work we will focus on homogeneous domain adaptation. The biggest challenge for this task is the different feature distributions between the domains, which is why most approaches map the domains to a common subspace \cite{Saito2018,Yao2019a}. Much research in this area focuses on unsupervised domain adaptation where no labels in the target domain are available \cite{Long2016,Saito2018,wang2021discriminative}. Matching the distributions of source and target features in a shared subspace is the key component of these methodologies and their main differentiating element. One line of research considers generative adversarial learning methods between the latent feature generator and the classifier to make the latent features of both domains indistinguishable  \cite{Saito2018,Ganin2017}. Other methods directly minimize some statistical distance measures between the domains like MMD \cite{Long2015,Long2016}. The authors of \cite{Xu2019} found that a stepwise adaption of feature norms of both domains results in significant transfer gains.  \\
Semi-supervised domain adaptation on the other hand assumes that some labeled examples of all classes are available in the target domain. The methodologies also rely mostly on an aligned shared latent feature space but are able to do a semantic alignment \cite{Motiian2017a} hence aligning the class conditional distributions. Similar to the UDA case many state of the art algorithms apply adverserial training \cite{Motiian2017,Saito2019}. 

\section{Methodology}
\begin{figure*}
    \centering
    \includegraphics[scale=0.3]{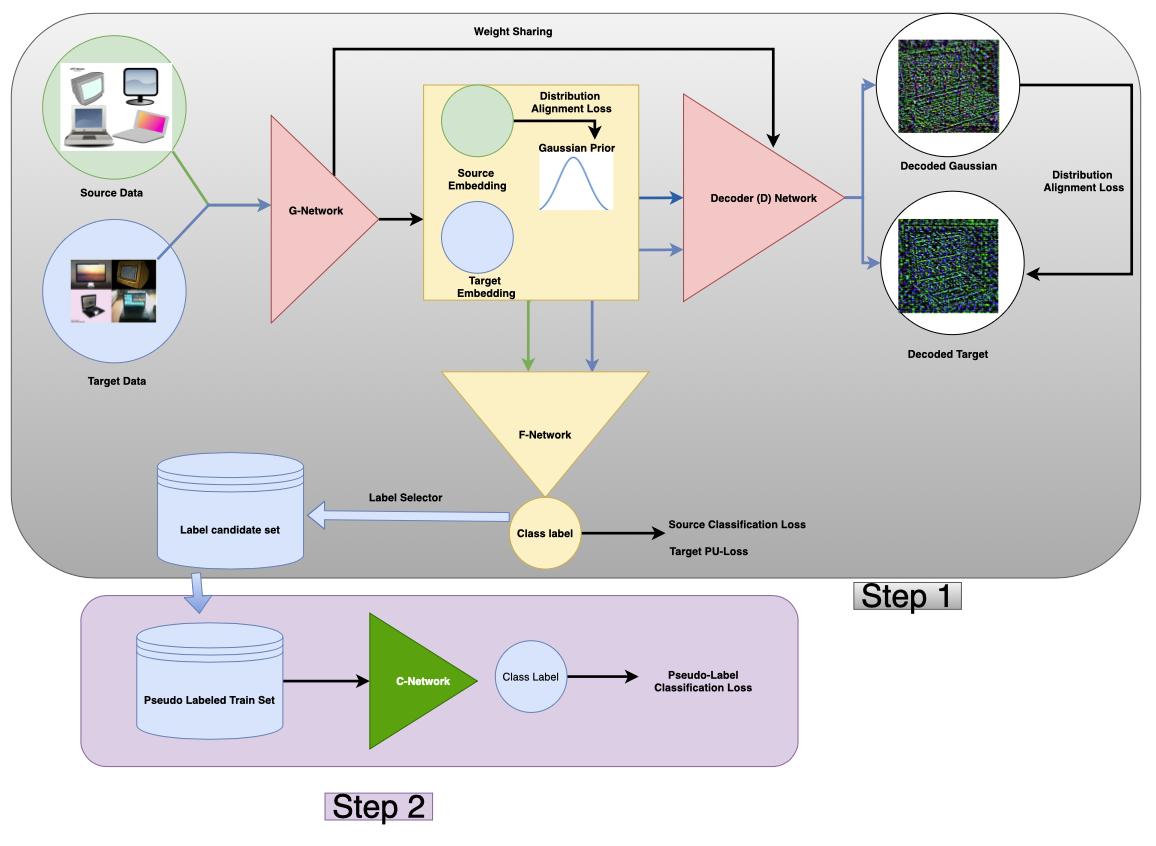}
    \caption{Overview of our model architecture (best viewed in colour). In the first step we identify a candidate set of likely positive and negative data based on  classical loss on source domain and PU loss on target domain after aligning the distribution space between both domains. In the second step we train a ResNet classifier on the pseudo-labeled target set. Only the classifier from the second step is used to do inference.}
    \label{fig:architecture}
\end{figure*}
Given a source domain $\mathcal{D}_S = \{(x_i,y_i)\}_{i=1}^{n_S}$ with $n_S$ labeled samples and a target domain $\mathcal{D}_T =\mathcal{D}_T^P \cup \mathcal{D}_T^U$ where $\mathcal{D}_T^P = \{(x_i,1)\}_{i=1}^{n_P}$ contains positive examples from the target domain and $\mathcal{D}_T^U = \{(x_i)\}_{i=1}^{n_U}$ are unlabeled examples from the target domain, containing both positive and negative examples. Usually we have $n_P \ll n_U$ and $n_P \ll n_S$. \\
We train on $D_S$, $D_T^P$ and $D_T^U$ and evaluate on $D_T^U$. \\
For our setting we focus on domain transfer between images of same resolution, so $\mathcal{D}_S$ and $\mathcal{D}_T $ live on the same feature space but with different feature distributions. \\
We propose a two step learning approach where in the first step we identify reliable negative and additional positive data in the target domain based on labeled source examples and positive target data using a PU risk estimator combined with an unsupervised domain adaptation algorithm. In the second step we then apply standard supervised learning models to the pseudo-labeled target data. Our algorithm is summarized in Algorithm \ref{alg:algorithm} and visualized in Figure \ref{fig:architecture}.
\subsection{Step 1 - Constructing pseudo target labels}
\subsubsection{Warm-Up Phase}
The key idea of our approach is to identify a subset of the unlabeled target data which we can label with high confidence based on the labeled source dataset. It is widely known that the naive transfer from a model between different datasets performs poorly due to feature distribution shift between the domains. In Domain Adaptation it is therefore common to learn a shared embedding space between the two domains with some (deep) network $G$ which is stacked together with a classifier $F$ such that the predicted class probability is calculated as $p(x) = softmax(F(G(x))$ where $x$ is either from $D_S$ or $D_T$.  
For our purpose we build upon the state-of-the art feature alignment methodology from \cite{wang2021discriminative}. The authors propose a discriminative feature alignment (DFA) using an indirect alignment of the shared embedding space guided by a Gaussian prior distribution together with an encoder-decoder architecture. Given a Gaussian prior distribution $z \sim \mathcal{N}(0,1)$ we follow their work and define two distribution alignment losses for source and target respectively.
The source domain is aligned to the Gaussian prior directly in the shared embedding space using Kullback-Leibler divergence $D_{KL}$. 
\begin{equation}
    \mathcal{L}_{S} = D_{KL}(z \ \Vert  \ \mathcal{D}_S). 
\end{equation}
The target domain is first reconstructed using the decoder network $D$ together with the prior distribution before aligning their distributions. According to the authors this ensures the learned embedding to be discriminative. The alignment loss on the target is therefore    
\begin{equation}
    \mathcal{L}_{T} = \frac{1}{n_T} \sum_{x \in \mathcal{D}_T} \lVert D(G(x))-D(z) \rVert_1,
\end{equation}
where $\lVert x\rVert_1$ is the L1-norm of $x$. 
The DFA algorithm is also leveraging the stepwise adaptive feature norm (SAFN) loss from \cite{Xu2019}. The authors found that including a regularization for the feature norm enlargement helps in transfer tasks since any domain gap is to some extend due to misaligned feature norm distributions. Denote for the classifier network $F$ with $l$ layers the first $l-1$ layers as $F_l$ and with $h(x) = \lVert F_l(G(x))\rVert_2$ the L2-norm of the feature representation in the last layer. The loss is defined as  
\begin{equation}
    \mathcal{L}_{SAFN} = \lVert h(x, \theta_P) + \delta - h(x, \theta_C) \rVert_2,
\end{equation}
where $\theta$ describes the network parameters of the current- ($\theta_C$) and the previous ($\theta_C$) training iteration with $\delta$ being a hyperparameter to control the feature norm enlargement in each step. \\

Putting it all together, we define the sum of the distribution alignment losses stemming from DFA as 
\begin{equation}
 \mathcal{L}_{align} =  \mathcal{L}_{S} + \mathcal{L}_{T} + \mathcal{L}_{SAFN}.  
\end{equation}

We further denote with 
\begin{equation}
\mathcal{L}_{cls} = \frac{1}{n_S} \sum_{(x_i,y_i) \in \mathcal{D}_S} y_i \log(p(x_i)) + (1-y_i) \log(1-p(x_i)) 
\end{equation}
the cross-entropy loss on the source domain. 
Since the DFA algorithm is designed for unlabeled domain adaptation and hence is not leveraging the positive labeled target data, we include an additional loss term $L_{PU}$ to the architecture which is the non-negative unbiased PU-risk estimator from \cite{Kiryo2017}. It leverages the fact that we can regard the unlabeled data as mixture of positive and negative examples which a prior probability $\pi$ of an example being positive, hence
\begin{equation}
    p(x) = \pi p(x |y=1) + (1-\pi) p(x|y=0). 
\end{equation}
The class prior $\pi$ can be estimated from unlabeled data \cite{Jain2016,Christoffel2016} and is assumed to be known throughout this paper. Applying their unbiased PU-loss formulation for a given loss function $l$ leads to   
\begin{align*}
\mathcal{L}_{PU} &= \frac{\pi }{n_P}\sum_{x \in \mathcal{D}_T^P} l(F(G(x)), 1) + \max \big{(}0,\\
& \quad \frac{1}{n_U}\sum_{x \in \mathcal{D}_T^U} l(F(G(x)), 0) -  \frac{\pi}{n_P}\sum_{x \in \mathcal{D}_T^P} l(F(G(x)), 0) \big{)}.
\end{align*}

Our final objective function is therefore
\begin{equation}
\label{eq:loss}
    \mathcal{L} = \mathcal{L}_{PU} + \alpha \mathcal{L}_{cls} + \beta \mathcal{L}_{align}.
\end{equation}
Since the PU-loss is the only measure defined on the target domain we consider it as most important and define regularization weights $\alpha$, $\beta$ for both other loss terms. \\

We train the network for 20 epochs in the warm-up phase minimizing \eqref{eq:loss} before we start to identify likely positive and negative target data using our label selection methodology. 

\subsubsection{Label Selector}
We apply the label selector each epoch $i$ after the warm up phase until the end of the first step. The label selector is selecting a subset $\mathcal{D}_L^i$ of $\mathcal{D}_T^U$ and stores it into a label-candidate set $D_{c}^i$. 
We make use of the positive labeled target examples to compute the threshold $T_{pos} = \frac{1}{n_P} \sum_{x \in D_T^P} p(x)$ and select those $x \in D_T^U$ as likely positive where $p(x)>T_{pos}$ holds. For the negative data we estimate the threshold using the source domain and define $T_{neg} = \frac{1}{n_S^N} \sum_{x \in D_S^N} p(x)$ where $D_S^N=\{(x, y) \in \mathcal{D}_S | y=0\}$. We select those $x \in D_T^U$ as likely negative where $p(x)<T_{neg}$ holds. We use the predicted positive class probabilities as pseudo label and obtain
\begin{equation}
   \mathcal{D}_c^i = \{(x, p(x)) \in \mathcal{D}_T^U \ | \ p(x)>T_{pos} \text{ or } p(x) < T_{neg} \}.  
\end{equation}
 Note that unlike other self-learning approaches like \cite{Chen2020} we do not remove label-candidates from the unlabeled dataset. This ensures that the PU-risk is estimated on all initially unlabeled data during the whole training period in step 1. Another implication is that the same observation can be added to $\mathcal{D}_c$ multiple times with different $p(x)$. We even expect a confident examples to be consistently classified with high confidence throughout the epochs. We therefore include a final label extractor logic that computes the pseudo-target-labels for a subset of $\mathcal{D}_c$ after the first step is finished, i.e. the model was trained for the specified number of epochs .
 Let $D_c = \bigcup_{i} D_{c}^i$ and
 % \begin{equation}
 %    |\cdot| = \sum_{x \in D_c} I(x=\cdot)  
 %\end{equation}
 \begin{equation}
     Count(x; \mathcal{D}_c) =  \sum_{(x',p) \in D_c} I(x=x').  
 \end{equation} 
 Then we construct the final pseudo-labeled dataset $\mathcal{D}_{pseudo} = \mathcal{D}_{pseudo}^P \cup \mathcal{D}_{pseudo}^N $ 
 as
 \begin{align}
 \begin{split}
 \label{eq:pseudo_p}
     \mathcal{D}_{pseudo}^P = \{(x, \overline{p(x)}) \in  \mathcal{D}_c \ | &\ Count(x) \geq m \text{ and } \\
     & \ \min(p(x)) > t_P \}
\end{split}
 \end{align}
 
 and 
 \begin{align}
 \begin{split}
 \label{eq:pseudo_n}
     \mathcal{D}_{pseudo}^N = \{(x, \overline{p(x)}) \in  \mathcal{D}_c \ | & \ Count(x) \geq m \text{ and } \\
     & \ \max(p(x)) < t_N \}
\end{split}
 \end{align}
 where $t_N$ and $t_P$ are thresholds for the minimum required negative and positive confidence respectively and $m$ is the minimal required number of occurrences of $x$ in  $\mathcal{D}_c$. We set those parameters empirically to $0.05$, $0.95$ and $20$, respectively, throughout all experiments. 
 
 \subsection{Step 2 - Obtaining the final classifier}
 After finishing step 1 and building $\mathcal{D}_{pseudo}$ we use this pseudo labeled dataset for training a classifier $C$ using a standard cross entropy loss 
\begin{align}
\begin{split}
\label{eq:step2}
\mathcal{L}_{cls} = \frac{1}{|D_{pseudo}|} & \sum_{(x_i,y_i) \in \mathcal{D}_{pseudo}} y_i \log(p(x_i) \\
 & + (1-y_i) \log(1-p(x_i)) 
 \end{split}
\end{align}
 We initialise $C$ with a pretrained ResNet-34 \cite{He2016} architecture trained on ImageNet, which is common technique in domain transfer \cite{Sankaranarayanan2018}, and train until a stopping criterion is reached. 

\begin{algorithm}[tb]
\caption{ PU-DA }
\label{alg:algorithm}
\textbf{Input}: $\mathcal{D}_S$, $\mathcal{D}_T^P$,  $\mathcal{D}_T^U$\\
\textbf{Output}: Trained Classifier $C$
\begin{algorithmic}[1] %[1] enables line numbers
\STATE Initialize $\mathcal{D}_{c} = \mathcal{D}_{pseudo} = \varnothing $ 
\STATE Initialize Networks $G$, $F$, $C$
\WHILE{$epoch<max\_epoch$}
\STATE Compute $T_{pos}$ and $T_{neg}$
\STATE Minimize \eqref{eq:loss} and update $G$, $F$
\IF {$epoch>warm\_up$}
\STATE $\mathcal{D}_{c} = \mathcal{D}_{c} \cup \{(x,p(x)) \in  \mathcal{D}_T^U \ | \ p(x) > T_{pos} \}$
\STATE $\mathcal{D}_{c} = \mathcal{D}_{c} \cup \{(x,p(x)) \in  \mathcal{D}_T^U \ |\ p(x) < T_{neg} \}$
\ENDIF
\ENDWHILE
\STATE Construct $\mathcal{D}_{pseudo}$ based on $\mathcal{D}_{c}$ using \eqref{eq:pseudo_p} and \eqref{eq:pseudo_n}
\WHILE{$epoch<max\_epoch$}
\FOR{$batch=1, \ldots N$}
\STATE Sample batch $(x_i, y_i)$ from $\mathcal{D}_{pseudo}$
\STATE Optimize \eqref{eq:step2} and Update $C$
\ENDFOR
\ENDWHILE
\end{algorithmic}
\end{algorithm}
\begin{table*}[]
\begin{center}
\begin{tabular}{lllllllll}
\hline
\textbf{Target}          & \textbf{c} & \textbf{DFA}   & \textbf{MME}   & \textbf{STN} & \textbf{FADA} & \textbf{NN-PU} & \textbf{PUbN} & \textbf{PU-DA(OURS)} \\ \hline
\multirow{3}{*}{Product} & 1          & $93.77^{0.8}$          & $\textbf{94.74}^{0.5}$          & $87.68^{1.6}$        & $85.53^{2.2}$         &       $61.12^{1.9}$         &    $58.59^{1.4}$           & $\textbf{94.91}^{1.5}$ \\ 
                         & 5          & $93.77^{0.8}$          & $\textit{94.02}^{0.4}$          & $92.35^{1.2}$        & $85.72^{2.3}$        &    $79.99^{4.5}$            &  $84.75^{0.8}$             & $\textbf{97.15}^{1.4}$ \\ 
                         & 10         & $93.77^{0.8}$          & $\textit{94.16}^{0.5}$          & $92.35^{2.5}$        & $85.16^{2.0}$         &       $91.57^{4.9}$         &   $93.4^{1.4}$            & $\textbf{97.58}^{1.2}$ \\ \hline
                         
\multirow{3}{*}{Art}     & 1          & $\textit{85.86}^{1.5}$          & $\textbf{87.38}^{0.8}$ & $64.37^{2.2}$        & $78.88^{2.4}$         &            $56.16^{1.2}$    &     $62.14 ^{0.2}$         & $79.04^{1.8}$           \\  
                         & 5          & $\textit{85.86}^{1.5}$        & $84.74^{0.8}$          & $73.29^{1.4}$        & $78.75^{2.9}$         &           $61.31^{5.0}$     &       $69.44^{1.3}$        & $\textbf{86.58}^{2.0}$ \\ 
                         & 10         & $\textit{85.86}^{1.5}$          & $83.60^{1.0}$          & $77.96^{1.1}$        & $78.70^{2.9}$         &           $64.14^{5.2}$     &      $76.82^{4.8}$         & $\textbf{87.09}^{1.7}$ \\ \hline
\multirow{3}{*}{Real}    & 1          & $\textbf{91.98}^{1.1}$  & $\textit{91.63}^{0.7}$          & $73.61^{2.8}$        & $84.91^{2.2}$         &              $60.31^{6.0}$  &   $73.37^{2.3}$            & $88.65^{1.5} $         \\ 
                         & 5          & $\textit{91.98}^{1.1}$          & $90.78^{0.8}$          & $86.67^{3.2}$        & $84.22^{2.5}$         &              $65.61^{7.2}$  &    $61.76^{3.7}$           & $\textbf{92.52}^{1.5}$ \\  
                         & 10         & $91.98 ^{1.1}$         & $\textbf{92.8}^{0.8}$  & $89.18^{2.0}$        & $82.69^{2.7}$        &     $73.45^{7.6}$           &  $90.87^{1.7}$             & $\textit{92.0}^{1.4}$           \\ \hline
\multirow{3}{*}{Clipart} & 1          & $\textit{87.71}^{1.8}$         & $\textbf{88.3}^{0.8}$  & $86.41^{4.4}$        & $81.99^{3.1}$         &              $62.06^{5.0}$  &       $68.57^{3.5}$        & $80.43 ^{2.6}$         \\ 
                         & 5          & $87.71^{1.8}$          & $86.98^{0.6}$          & $\textit{88.93}^{2.6}$       & $84.25^{1.9}$         &         $70.22^{5.0}$       &    $74.87^{1.7}$           & $\textbf{90.81}^{2.7}$ \\  
                         & 10         & $87.71^{1.8}$          & $86.36^{0.6}$          & $\textit{89.65}^{1.9}$        & $84.23^{1.6}$         &            $77.64^{4.9}$    &     $86.01^{0.3}$          & $\textbf{91.95}^{1.8}$ \\ \hline \hline
\multicolumn{2}{l}{\textbf{Summary Score}}  & 3.5 & 5.5 &1 & 0 & 0&0 & \textbf{8.5}
                         
\end{tabular}
\end{center}
\caption{\label{tab:office-home}Accuracy and standard deviation on unlabeled target data for Monitor vs. Laptop classification on Office-Home dataset. \\ All results are averaged over 20 runs and three adaptation scenarios. We compute a summary score per methodology to allow an easy performance comparison across different domains and label frequencies.  We award one point for the best performing model and half a point for the second best model per scenario.}

\end{table*}

\section{Experiments}
We validate our methodology on well known benchmark object recognition and image classification datasets for domain adaptation, namely Office-Home \cite{venkateswara2017deep}, Office-Caltech \cite{Saenko2010} and MNIST-USPS \cite{Hull1994} which are all publicly accessible. We binarize the prediction problem by choosing two visually similar classes as positive and negative respectively while dropping all other classes. In the target domain we remove labels according to the label frequency $c = P(s=1 | y=1)$, where $s$ is the property of an example being labeled. Hence we keep $c$ percent of the positive labels and consider everything else as unknown. We conduct every experiment for label frequencies 1, 5 and 10 percent. \\  
All results are reported as the average accuracy together with standard deviation of twenty training runs, where each training uses a different randomly selected labeled subset in the target domain. The standard deviation is therefore mainly measuring the influence of different labeled positive subsets and not so much the influence of stochasticity in the learning optimization process itself. \\  
We perform hyperparameter training for each (baseline-) model using baysian search \cite{Perrone2019}. We report all hyperparameters and their chosen values in the supplementary material. 

\subsection{Baselines}
We consider state of the art baselines from unsupervised- and semi-supervised Domain Adaptation as well as from PU learning. All baselines are implemented based on their public available code repositories with small adaptions related to the structure of our datasets which is described in more detail in the supplementary material.  \\

\textbf{DFA} \cite{wang2021discriminative} denotes the discriminative feature alignment methodology by a gaussian prior which we introduced in the methodology section as basis for the domain alignment we use. Since this algorithm performs unsupervised domain adaptation, this is our baseline for considering only the domain adaptation problem, ignoring the positive labels in the target domain.  \\
From semi-supervised domain adaptation methodologies we consider MiniMax Entropy (\textbf{MME})\cite{Saito2018}, Soft Transfer Network(\textbf{STN})\cite{Yao2019a} and Few-Shot Adversarial Domain Adaptation (\textbf{FADA})\cite{Motiian2017} methodologies. All of them provide strong performance under the assumption that few labels of every class are available in the target domain. Since these algorithms originally use examples of all classes we had to make small adaptations to apply them to our novel setting. STN computes a classwise loss in the target domain, so we removed the summand of the negative class from the loss function. FADA also requires an explicit negative class in the target domain which is used in a more complex way than simply as summand in the loss function, so we regarded all unlabeled data as negative. The MME algorithm uses all available target labels without explicitly separating the classes, so we were able to use it's original implementation.  \\
From PU research we consider \textbf{NN-PU}\cite{Kiryo2017} and \textbf{PUbN}\cite{Hsieh2019}. Both use special risk estimators to account for the PU setting. All PU baselines were applied to the target domain only since they are not designed to leverage domain transfer information. 

\subsection{Office-Home Task}
%\begin{figure}
%\centering
%\begin{subfigure}{.3\textwidth}
 % \centering
 % \includegraphics[width=.5\linewidth]{Real.png}
 % \caption{Real domain}
 % \label{fig:sfig1}
%\end{subfigure}%
%\begin{subfigure}{.3\textwidth}
%  \centering
%  \includegraphics[width=.5\linewidth]{Art.png}
 % \caption{Art domain}
 % \label{fig:sfig2}
%\end{subfigure}
%\begin{subfigure}{.3\textwidth}
 % \centering
%  \includegraphics[width=.5\linewidth]{Clipart.png}
 % \caption{Clipart domain}
%  \label{fig:sfig21}
%\end{subfigure}
%\begin{subfigure}{.3\textwidth}
 % \centering
 % \includegraphics[width=.5\linewidth]{Product.png}
 % \caption{Product domain}
 % \label{fig:sfig22}
%\end{subfigure}
%\caption{\label{fig:oh-example}Examples of positive monitor class (first row} and negative laptop class (second row) per domain in the Office-Home dataset. 

%\end{figure}

    \begin{figure}
        \centering
        \begin{subfigure}[b]{0.23\textwidth}
            \centering
            \includegraphics[width=0.7\textwidth]{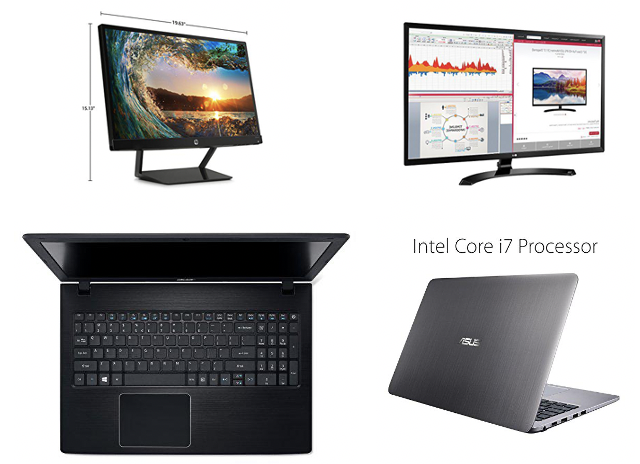}
            \caption[Product Domain]%
            {{\small Product Domain}}    
            \label{fig:mean and std of net14}
        \end{subfigure}
        \hfill
        \begin{subfigure}[b]{0.23\textwidth}  
            \centering 
            \includegraphics[width=0.7\textwidth]{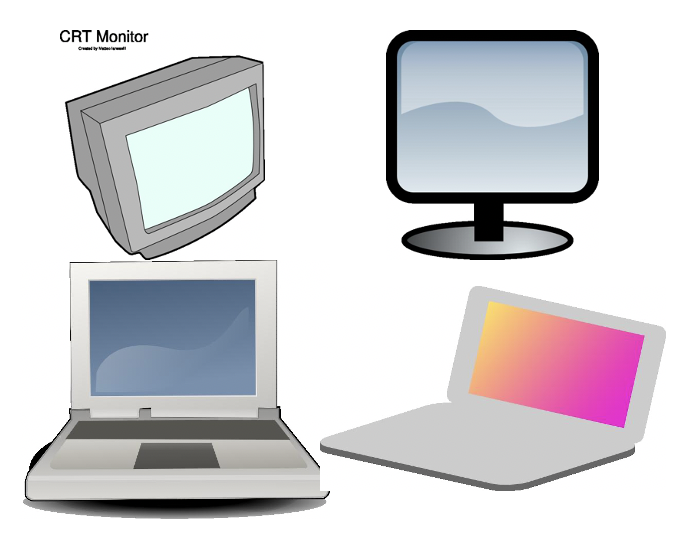}
            \caption[]%
            {{\small Clipart Domain}}    
            \label{fig:mean and std of net24}
        \end{subfigure}
        \vskip\baselineskip
        \begin{subfigure}[b]{0.23\textwidth}   
            \centering 
            \includegraphics[width=0.7\textwidth]{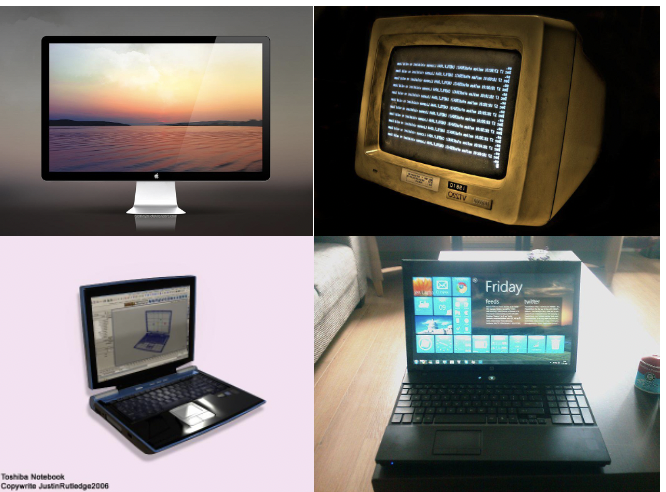}
            \caption[]%
            {{\small Art Domain}}    
            \label{fig:mean and std of net34}
        \end{subfigure}
        \hfill
        \begin{subfigure}[b]{0.23\textwidth}   
            \centering 
            \includegraphics[width=0.7\textwidth]{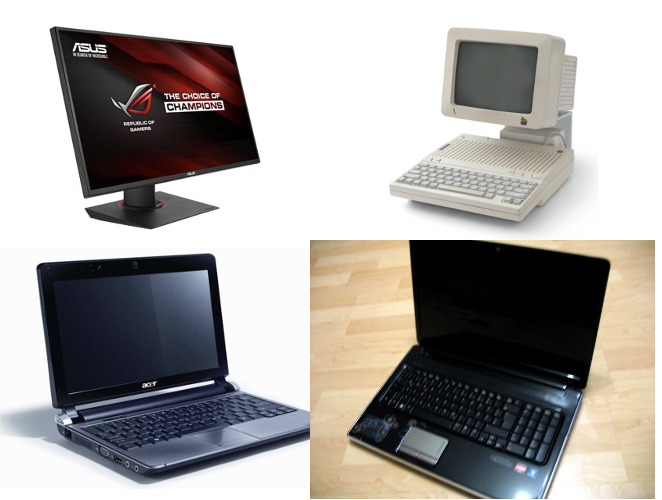}
            \caption[]%
            {{\small Real World Domain}}    
            \label{fig:mean and std of net44}
        \end{subfigure}
        \caption[ Examples of positive monitor class (upper row) and negative laptop class (lower row) per domain in Office-Home ]
        {\small Examples of positive monitor class (upper row) and negative laptop class (lower row) per domain in Office-Home  } 
        \label{fig:oh-example}
    \end{figure}
    
\begin{table*}[]
\begin{center}
\begin{tabular}{lllllllll}

 \hline
\textbf{Source}          & \textbf{c} & \textbf{DFA} & \textbf{MME} & \textbf{STN} & \textbf{FADA} & \textbf{NN-PU} & \textbf{PUbN} & \textbf{PU-DA(OURS)} \\ \hline
\multirow{4}{*}{Webcam}  & 1          & $91.57^{2.5}$        &    $\textbf{94.96}^{0.5}$         &    $91.10^{2.7}$  &     $79.56^{6.1}$   &          $66.61^{7.0}$    &     $84.66^{1.7}$            & $\textit{94.35}^{3.2}$              \\
                         & 5          & $89.47^{2.5}$        &    $\textit{93.63}^{0.4}$        &     $93.5^{1.4}$  &      $79.41^{4.3}$ &          $74.09^{4.8}$        &    $85.03^{1.1}$         & $\textbf{94.76}^{2.7}$            \\
                         & 10         & $89.47^{2.5}$        &  $93.34^{0.5}$            &     $94.30^{1.3}$  &     $82.17^{2.5}$  &          $88.88^{1.0}$     &     $\textbf{94.38}^{0.5}$           & $\textbf{95.15}^{2.2}$                \\
                        % & 20         & 91.57        &   92.49           &      &        &               &                & 93.93                \\                           \hline
 \multirow{4}{*}{Caltech} & 1          & $91.57^{2.0}$        &   $\textit{94.27}^{0.6}$            &  $84.90^{1.5}$     &      $90.70^{1.7}$ &             $66.61^{7.0}$  &  $84.66^{1.7}$             & $\textbf{95.06}^{0.6}$               \\ 
                         & 5          & $91.57^{2.0}$        &   $93.53^{0.4}$            &    $\textit{94.31}^{0.9}$    &    $88.97^{2.0}$  &               $74.09^{4.8}$  &   $85.03^{1.1}$           & $\textbf{94.81}^{0.6}$                \\
                         & 10         & $91.57^{2.0}$        &    $92.81^{0.6}$          &     $94.35^{0.6}$   &    $89.19^{1.3}$  &              $88.88^{1.0}$   &     $\textit{94.38}^{0.5}$         & $\textbf{95.10}^{1.0}$                \\
                         % & 20         & 89.47        &   91.51           &              &               &                & 94.35                \\ 
                          \hline  \hline
 \multicolumn{2}{l}{\textbf{Summary Score}}  & 0 & 2 &0.5 & 0 & 0&1.5 & \textbf{5.5}
 \end{tabular}
 \end{center}
 \caption{\label{tab:results_office}Accuracy and standard deviation on the Office-Caltech classification task for target amazon}
 \end{table*}    
The Office-Home dataset contains 4 domains (Real (R), Clipart (C), Art (A), Product (P)) with 65 classes. To make the binary prediction class challenging we consider two visually similar classes as positive and negative, namely computer-monitor and laptop. Figure \ref{fig:oh-example} shows example images of each domain. One special property of this dataset is that the number of images per class is rather limited. We have approximately 100 images per class which makes the label percentage $c$ equal to the absolute number of labeled target examples. \\
The four domains allow 12 adaptation scenarios in total and we evaluate our methodology on all of them. For each adaptation scenario we investigated the influence of the labeled prior in the target domain which leads to a total of 36 experiments. For simplicity we report the averaged result per target in Table \ref{tab:office-home}, e.g. the results for target product is the averaged result of three adaptation scenarios (R $\rightarrow$ P, C $\rightarrow$ P, A $\rightarrow$ P). We provide the full results in the supplementary material. \\
We compute a summary score per method to allow a quick overall comparison of methodologies over different adaptation scenarios and labeled priors. The summary score is obtained by awarding one point for each scenario where the methodology performs best and half a point for the second best methodology. If there is no statistically significant difference (two-sided T-test, p-value smaller 5 percent) between the best methodologies we provide one point for each of them. We find that our approach has the best performance overall, followed by SSDA approach MME and UDA approach DFA. We therefore can conclude that among the SSDA baselines MME can best leverage information from a PU subset in the target domain, while FADA is by design not performing well in our setting. Also we find the PU-approaches to perform worse than DA which is indicating that the source domain contains more information than the small labeled positive subset in the target domain. The strong performance of our methodology shows however that combining both sources of information is beneficial. \\
Taking a closer look at the single scenarios we note that our methodology is mainly outperforming all baselines for 5- and 10-shot learning of positive labels in the target domain. However we find that for the one-shot scenario our methodology seems to perform worse than unsupervised domain adaptation. Since the labeled positive data is guiding our selection process of new trusted positives, a single, unrepresentative labeled target data apparently hurts the performance. We plan to investigate how to make the approach more robust in the one-shot scenario in future works. \\ 
We also find that even though MME is mostly outperforming unsupervised domain adaptation, increasing the number of labeled target data seem to even hurt model performance of MME since it introduces a bias towards positive predictions. \\
Although not surprising, it is worth mentioning that the performance gap between DA and PU approaches is getting smaller with an increasing number of positive labels in the target domain. \\
With respect to standard deviation we find that especially the NN methodologies varies quite much in its performance. This is indicating that in few shot scenarios pure PU-learning performance highly depends on the representativity of the labeled subset. Methodologies that also use the source domain are more robust in their performance.

\subsection{Office-Caltech Task}
The Office dataset contains pictures of different product categories from three distinct sources (Webcam, DSLR and Amazon) which differ in resolution and background noise. The Caltech-256 dataset shares ten product categories with the office dataset and therefore they are often combined for domain adaptation tasks \cite{Gong2012}. We follow \cite{Yao2019a} and consider amazon data as target domain and investigate caltech and webcam as sources respectively. While amazon and caltech are of similar size with little more than 100 examples per class, webcam is even smaller providing only 43 positive and 30 negative examples.      
We apply the same setting as in Office-Home and investigate the scenario computer-monitor vs laptop. \\
In terms of overall performance we again find that our methodologies outperforms all baselines from which MME again is second-best.   
Looking at the single scenarios, we find that unlike in the Office-Home dataset there is no significant difference in performance of our algorithm between different labeled priors. Since the amazon domain contains images taken from amazon marketplace which are usually of high quality and similar to each other within one class, even one randomly selected image seem to be representative for the positive domain and enables our algorithm to build a meaningful pseudo-labeled dataset. \\ 
We also find that 10 labelled positive examples are enough for the PUbN methodology to outperform domain adaptation, i.e. for the amazon data 10 available labels in the target domain enable a better prediction than the fully labeled source domain. This again is an indication that the classes are better separated in the amazon domain compared to the Office-Home domains.  
This hypotheses is in line with the high performance in the product domain of the Office-Home task since the product domain like the amazon dataset contains pictures of products sold in some online marketplace and are therefore less noisy than domains like art or clipart. \\
We therefore find that our methodology performs strongly given a representative positive labeled subset. \\
We find that over both experiments only 5 labeled datapoints in the target are sufficient for us to consistently outperform all baselines. Whether our methodology works on a single labeled datapoint is dependent on the domain, where we outperform baselines in domains of high picture quality and class separability like amazon or product. 

\subsection{Digit Task}
\begin{table}[]
\begin{tabular}{|l|lll||l|}
\hline
\tikz{\node[below left, inner sep=1pt] (domain) {Domain};%
       \node[above right,inner sep=1pt] (c) {c};%
       \draw (c.north west|-domain.north west) -- (domain.south east-|c.south east)}                                                      & \textbf{1}     & \textbf{5}     & \textbf{10}    & \textbf{\begin{tabular}[c]{@{}l@{}}Summary \\ Score\end{tabular}} \\ \hline
\textbf{DFA}                                                    & $\textit{96.09}^{2.4}$ & $96.09^{2.4}$          & $96.09^{2.4}$          & 0.5                                                               \\ \hline
\textbf{MME}                                                    & $93.41^{0.6}$          & $95.94^{0.6}$          & $\textit{96.79}^{0.7}$ & 0.5                                                               \\ 
\textbf{STN}                                                    & $92.1^{1.0}$           & $95.6^{1.2}$           & $96.5^{0.9}$           & 0                                                                 \\ 
\textbf{FADA}                                                   & $49.92^{2.3}$          & $60.68^{6.9}$          & $61.73^{5.7}$          & 0                                                                 \\ 
\textbf{NN-PU}                                                  & $52.25^{6.8}$          & $72.15^{1.3}$          & $84.22^{1.1}$         & 0                                                                 \\ 
\textbf{PUbN}                                                   & $92.68^{2.5}$          & $\textit{96.38}^{1.7}$ & $95.37^{1.7}$          & 0.5                                                               \\ 
\textbf{\begin{tabular}[c]{@{}l@{}}PU-DA\\ (Ours)\end{tabular}} & $\textbf{97.48}^{0.9}$ & $\textbf{97.2}^{1.1}$  & $\textbf{97.46}^{0.9}$ & 3                                                                 \\ \hline
\end{tabular}
\caption{\label{tab:digit_res} Accuracy and standard deviation on the adaptation scenario from MNIST to USPS for 3 vs 5 classification.}
\end{table}
Another commonly used benchmark for domain adaptation is the transfer between the digit datasets MNIST \cite{LeCun1998} and USPS \cite{Hull1994}. Both contain images of handwritten digits and have around 5000 training points per class. We follow \cite{Hou2018} and binarize the classification by considering the visually similar digits 3 as positive and 5 as negative class. We consider this dataset to investigate our methodology in a scenario where we have more labeled data in the target domain, i.e. even a small labeled prior of 1 percent is considering fifty labeled target images.
The results are summarized in Table \ref{tab:digit_res}. We clearly see that our approach is outperforming the baselines consistently over all scenarios. Similar to Office-Caltech we do not note significant differences between the different labeled priors which again indicates that a representative labeled subset even of small size allows to build a high quality pseudo-labeled set.    
\subsection{Real World Applications}
In the previous experiments we showed that our methodology outperforms baselines from PU learning and domain adaptation on well known benchmark datasets. However for the considered data it is unrealistic to assume that only positive lables are available in one domain while the other is completely labeled. We therefore like to discuss real world applications of our methodology and validate our methodology on a dataset from the mobility domain, namely parking availability prediction. 
\subsubsection{Parking Availability}
Predicting the availability of on-street parking locations is a relevant problem for traffic management in smart cities. Parking behaviour in different cities is related but suffers from domain shift \cite{Shao2021}, e.g. business areas tend to be occupied during the day but available in the evening in most cities while the distribution of those areas varies between cities. Many researchers are aiming at predicting the binary occupancy status of parking location where the location consists of several individual parking bays (e.g. all on-street parking bays between two intersections). A location is considered occupied if all individual parking bays are occupied, otherwise available \cite{Sonntag2021}. Some cities like Melbourne in Australia have installed in-ground sensors that detect the occupancy status of the parking bay\footnote{https://www.melbourne.vic.gov.au/about-council/governance-transparency/open-data/Pages/on-street-parking-data.aspx}. However due to the high costs of installing and maintaining such sensors only 5 percent of all individual parking bays in Melbourne are equipped with such a sensor. As a result negative availability labels can only be inferred for locations where every single bay is equipped with a sensor. Positive labels however are much easier to obtain since it is enough to observe just one available individual bay to label the whole location as available. To regard parking prediction as PU learning problem was first proposed by \cite{Sonntag2021}. \\
Our methodology can therefore be used to predict parking availability in cities which are only partially covered by sensors guided by fully sensorized cities. \\
We used those streets in Melbourne which are fully sensorized as source domain and as target we consider Seattle, US which also publishes its parking data\footnote{https://data.seattle.gov/Transportation/Annual-Parking-Study-Data/7jzm-ucez}. Details on the data processing and the considered features are provided in the supplementary material. Since the labels are imbalanced towards the positive class we report the balanced accuracy score \cite{Brodersen2010}. 
\begin{table}[]
\begin{tabular}{|l|lll||l|}
\hline
\tikz{\node[below left, inner sep=1pt] (domain) {Model};%
       \node[above right,inner sep=1pt] (c) {c};%
       \draw (c.north west|-domain.north west) -- (domain.south east-|c.south east)}                                                      & \textbf{1}     & \textbf{5}     & \textbf{10}    & \textbf{\begin{tabular}[c]{@{}l@{}}Summary \\ Score\end{tabular}} \\ \hline
\textbf{DFA}                                                    & $61.98^{1.0}$ &    $63.35^{1.3} $   &    $62.14^{1.0} $    &    0                                                            \\ 
\textbf{MME}                                                    & $62.36^{1.1}$ &    $62.36^{1.2}$   &   $62.37^{0.9}$    &    0                                                         \\ 
\textbf{STN}                                                     & $63.20^{0.7}$ &    $63.21^{0.7}$    &   $63.25^{0.8}$    &     0                                                  \\ 
\textbf{FADA}                                                   & $\textit{63.68}^{1.9}$ &   $\textit{63.67}^{2.0}$     &   $\textit{63.68}^{1.6}$     &    1.5                                                             \\ 
\textbf{NN-PU}                                                 & $61.13^{1.2}$ &     $61.11^{1.4}$   &    $61.13^{1.3}$   &     0                                                 \\ 
\textbf{PUbN}                                                  & $62.40^{1.7}$ &   $62.41^{1.7}$     &    $62.51^{1.4}$   &     0                                                    \\ 
\textbf{\begin{tabular}[c]{@{}l@{}}PU-DA\\ (Ours)\end{tabular}}  & $\textbf{64.84}^{1.7}$ &    $\textbf{65.33}^{1.5}$     &   $\textbf{65.09}^{1.5}$     &   3                                                     \\ \hline
\end{tabular}
\caption{\label{tab:parking_res} Balanced Accuracy and standard deviation on the parking availability scenario from Melbourne to Seattle.}
\end{table}
The results are summarized in Table \ref{tab:parking_res} and we clearly see our methodology outperforming all baselines. 
\subsubsection{Classification of Alzheimer Disease from MRI Images}
Image classification for medical MRI images suffer from domain shift since different scanners, scanning parameters and subject cohorts induce different feature distributions \cite{Guan2021}.
The classification of Alzheimer patients from MRI images on the other hand is an accepted baseline for PU learning\cite{Chen2020} since ill patients go through a transition period where they are seldom diagnosed. In this phase they are considered ill but are often labeled wrongly as healthy (negative). The authors in \cite{Chen2020} successfully considered all non-diagnosed patients as unlabeled instead of healthy. \\  
It is however in general possible to provide full labels in retrospective. This effort is much better invested when the labeled dataset can be used together with images on all kinds of scanners and scanner parameters. We therefore think it would be highly interesting to apply our methodology to medical image classification in future research. 

\section{Conclusion}
In this work we introduced a novel domain adaptation scenario were we aim to generalize from a fully labeled source domain to a target domain with only few positive and unlabeled data. Since there are naturally no methodologies designed for this setting yet, we proposed a relatively simple yet effective two-step learning approach inspired by recent work from semi-supervised learning and unsupervised domain adaptation. We thereby set the first baseline for positive-unlabeled domain adaptation. We benchmarked our approach on various publicly available datasets from image classification and found that our approach outperforms the baselines from (semi-supervised)-domain adaptation and PU-learning even in few shot scenarios with 5 and 10 positive target examples. \\
We also proposed real-world applications for our novel setting and validated our methodology on the use case of parking availability predictions. \\  
In future works we plan to develop further methodologies in our new setting inspired by approaches that are commonly used in DA- and PU-learning like GANs.   
\bibliographystyle{plain}
\bibliography{trasnfer_few_shot, PU_bib, Predparking}

\end{document}